\documentclass[preprint]{elsarticle}

\usepackage[english]{babel}
\usepackage[utf8x]{inputenc}
\usepackage[T1]{fontenc}
\usepackage{textcomp}
\RequirePackage[table]{xcolor}
\usepackage{collcell}
\usepackage{booktabs}
\usepackage{hhline}
\usepackage{pgf}
\usepackage{placeins}
\usepackage{amsfonts}
\usepackage{graphicx}
\usepackage[colorinlistoftodos]{todonotes}
\usepackage{lineno,hyperref}
\usepackage{siunitx}
\usepackage[onehalfspacing]{setspace}
\usepackage{bm}
\usepackage{array}
\newcolumntype{M}[1]{>{\centering\arraybackslash}m{#1}}
\usepackage{subcaption}
\usepackage{amsmath}

\usepackage{lipsum}
\usepackage{fancyhdr}

\newcommand\gray{gray}

\newcommand\ColCell[1]{%
	\pgfmathparse{#1<.8?1:0}%
	\ifnum\pgfmathresult=0\relax\color{white}\fi
	\pgfmathparse{1-#1}%
	\expandafter\cellcolor\expandafter[%
	\expandafter\gray\expandafter]\expandafter{\pgfmathresult}#1}

\newcolumntype{E}{>{\collectcell\ColCell}c<{\endcollectcell}}
\usepackage[inline]{enumitem}
\modulolinenumbers[5]

\journal{Robotics and Autonomous Systems}

\makeatletter
\def\@author#1{\g@addto@macro\elsauthors{\normalsize%
		\def\baselinestretch{1}%
		\upshape\authorsep#1\unskip\textsuperscript{%
			\ifx\@fnmark\@empty\else\unskip\sep\@fnmark\let\sep=,\fi
			\ifx\@corref\@empty\else\unskip\sep\@corref\let\sep=,\fi
		}%
		\def\authorsep{\unskip,\space}%
		\global\let\@fnmark\@empty
		\global\let\@corref\@empty  
		\global\let\sep\@empty}%
	\@eadauthor={#1}
}
\makeatother

\begin{document}

	\begin{frontmatter}

		\title{Robust and Subject-Independent Driving Manoeuvre Anticipation through Domain-Adversarial Recurrent Neural Networks}

		\author[add1]{Michele~Tonutti\corref{cor1}}
		\ead{michele.tonutti@pacmed.nl}
		\author[add2]{Emanuele~Ruffaldi}
		\author[add3]{Alessandro~Cattaneo}
		\author[add4]{Carlo~Alberto~Avizzano}

		\cortext[cor1]{Corresponding author}
		\address[add1]{Pacmed, Amsterdam, The Netherlands}
    \address[add2]{Medical Micro Instruments, Calci (PI), Italy}
		\address[add3]{Teoresi Group, Assago (MI), Italy}
		\address[add4]{PERCRO Laboratory, Scuola Superiore Sant'Anna, Ghezzano (PI), Italy}

		\begin{abstract}
		Through deep learning and computer vision techniques, driving manoeuvres can be predicted accurately a few seconds in advance. Even though adapting a learned model to new drivers and different vehicles is key for robust driver-assistance systems, this problem has received little attention so far. This work proposes to tackle this challenge through domain adaptation, a technique closely related to transfer learning. A proof of concept for the application of a Domain-Adversarial Recurrent Neural Network (DA-RNN) to multi-modal time series driving data is presented, in which domain-invariant features are learned by maximizing the loss of an auxiliary domain classifier. Our implementation is evaluated using a leave-one-driver-out approach on individual drivers from the Brain4Cars dataset, as well as using a new dataset acquired through driving simulations, yielding an average increase in performance of 30\% and 114\% respectively compared to no adaptation. We also show the importance of fine-tuning sections of the network to optimise the extraction of domain-independent features. The results demonstrate the applicability of the approach to driver-assistance systems as well as training and simulation environments.
		\end{abstract}


	\end{frontmatter}

	\section{Introduction}

	With 1.3 million deaths and 30 million injuries occurring yearly worldwide, road traffic accidents are the main cause of death for people aged 15-29, and represent a cost to governments of, on average, 3\% of national GDPs \citep{Paul2016}. A high proportion of those accidents occur during manoeuvres such as changing lanes and turning \citep{Hurt1981}. Advanced Driver Assistance Systems (ADAS) aim at increasing road safety by taking partial control of the car or by providing the driver with extra information when such manoeuvres could be dangerous \citep{Levinson2011TowardsAlgorithm}. It has been demonstrated that, thanks to recent developments in deep learning and computer vision, it is possible to predict manoeuvres a few seconds in advance and with high accuracy, by monitoring the driver's behaviour inside the vehicle and using information from the car itself (e.g. speed) and the environment (lanes configuration, presence of intersections, etc.). In particular, advances in Convolutional Neural Networks (CNN) now allow accurate extraction of head-, face-, and gaze-related features from videos \citep{Le2011,Karpathy2014}, while Recurrent Neural Networks (RNN) enable the models to take into account the temporality of an action, i.e. the order in which certain actions are performed or specific events occur \citep{Hermans2013}.\\
	However, while most recent proof-of-concept models for manoeuvre anticipation have achieved good results, little attention has been given to the problems arising from the practical implementation of such systems. One of the main concerns is their ability to generalise on subjects that were not part of the original training set. In real world applications, it is likely that ADAS installed on commercial cars will be "blind" to the driving style of new subjects. Re-training such systems may not feasible due to the lack of labelled examples for a new driver. While there are examples of deep neural networks trained to anticipate actions and objects in videos through unsupervised learning \citep{Vondrick2015}, most of the examples found in literature are not able to learn temporal relationships between features, or work exclusively with video inputs \citep{Ravichandar2016b}. This is a problem in manoeuvre anticipation tasks, since it has been demonstrated that multi-modal inputs and the temporality of events are crucial to obtain quick and accurate predictions. Moreover, if they are not re-trained, classic deep neural networks tend to not generalize well when features in the test and training sets have different marginal distributions. Examples could include cases in which the driver has peculiar driving habits or mobility limitations, so that, for instance, they cannot turn their head fully. It is however plausible to assume that there exist common patterns and latent features in the actions of most drivers which are common regardless of the vehicle, driving style, or situation \citep{Ben-David2007}.\\
	This assumption represents the basis of domain adaptation, a type of transfer learning technique which enables previously trained models to adapt to other datasets containing unlabelled observations \citep{Ben-David2010}. Amongst the various approaches, domain adaptation applied to deep neural networks has been proven to be extremely effective to learn domain-invariant features from the input data even when the labels of the target distribution are unknown \citep{Ganin2016Domain-adversarialNetworks}. This enticing result is obtained by optimising a discriminative classifier while simultaneously maximising the loss of an auxiliary domain classifier. This technique has been proven to work not only with image inputs, but also with time-series observations \citep{Purushotham2016a}. Using this method, we hypothesize that a model can be trained to find features in sequential input data which are not only discriminative of specific manoeuvres, but also shared between the training set (large and labelled) and a fully or mostly unlabelled small test set - such as video segments of a new driver performing unknown manoeuvres. Finding such latent features may, however, be difficult when the inputs are multi-modal time-series. This type of input is common in state-of-the-art manoeuvre anticipation models, which integrate the driver's behaviour with information from the vehicle and the environment - such as lane configuration, car speed, and GPS data \citep{Jain2016}. The network does not only need to learn the dependencies \textit{within} a single time-series, but also \textit{across} input sequences with possibly very different resolution (sparse vs. dense) or even data types (e.g. categorical vs. continuous). An effective sensory fusion approach is therefore required.

	\subsection{Aim}
	Given the problem of adaptation and generalization capabilities of multi-modal models for driver-assistance systems, we present an investigation of the application of the domain-adversarial training method to implement domain adaptation on a Recurrent Neural Network for manoeuvre anticipation. More generally, this paper proposes domain adaptation as a promising approach to improve the performance and generalization ability of machine-learning-driven ADAS. We aim at showing that domain-adversarial models are particularly beneficial in situations where either the drivers or the driving settings –or both– may differ considerably to those used to train the model.

	\subsection{Method and Structure}
	 In order to evaluate our approach, we designed three experiments, which are introduced in the following paragraphs and discussed more in depth in Section \ref{sec:experiments}. Before presenting the results of the experimental work, we provide a bibliographical review of the most recent published work on action prediction, maneuver anticipation, and domain adaptation. We then present a technical overview on Recurrent Neural Networks. Finally, we describe the architecture of our models and the most salient features of both the Brain4Cars and our own dataset, including a detailed explanation of the data collection process.
	 The experimental set-ups consist of the following:

	 \textbf{Experiment 1}
	 We first propose an expansion of the architecture proposed by Jain et al. \citep{Jain2016} (Brain4Cars), which includes the driver's gaze as an additional input, features a higher number of stacked recurrent layers, and performs enhanced sensory fusion by combining Long Short Term Memory (LSTM) and Gated Recurrent Unit (GRU) layers, amongst other improvements. We replicate the experiments presented in the original paper, training and testing the LSTM-GRU model on the Brain4Cars dataset using simple 5-fold cross-validation. This experiment is aimed at showing that the performance of our architecture is comparable to state-of-the-art models for maneuver anticipation in non-adaptive tasks. Additionally, we also test the model on a new set of driving videos obtained using an immersive virtual simulation setup, providing a performance baseline upon which to evaluate the results of the subsequent experiments.

	 \textbf{Experiment 2}
	 We then present a Domain-Adversarial RNN, inspired by Ganin et al. \citep{Ganin2016Domain-adversarialNetworks}, in which our LSTM-GRU network from Experiment 1 serves as the feature extractor. We train and test the DA-RNN using a cross-validated, leave-one-driver-out approach on individual drivers from the Brain4Cars dataset, comparing its performance to the non-domain-adaptive LSTM-GRU trained without the target driver.

	 \textbf{Experiment 3}
	 Finally, we implement the same domain-adversarial approach to study how the network, trained only on the Brain4Cars data, adapts to our new dataset, in which the drivers and the driving set-up –e.g. position of the mirrors, windows, and the camera– differ from the Brain4Cars dataset.

	 The results confirm that, without adaptation, the model is not able to predict manoeuvres from observations in which the features have very different marginal distributions compared to the training set. We conclude discussing the potential applications of the domain-adversarial approach to apply domain adaptation in commercial ADAS, driving training set-ups, and simulation environments. \\

	 The crucial contributions of this work can thus be summarised as follows:
	\begin{itemize}
		\item An improved LSTM-GRU Neural Network architecture for manoeuvre anticipation.
        \item An original dataset of observations from a driving simulation setup.
		\item A proof of concept for the application of a Domain-Adversarial RNN for domain adaptation on driving data, which employs the LSTM-GRU architecture to classify observations from the new dataset.

	\end{itemize}

\section{Related Work}
	\label{sec:related_work}

	The majority of studies relying on driving data do not concern themselves with the adaptation of the system to different datasets or their ability to perform well on new drivers; those who aim at anticipating manoeuvres are not an exception. Samples from all test subjects are often shuffled together before the training-test split \citep{Jain2016}: while this enhances a classifier's performance, it will likely cause it not to generalise and scale well in real-world applications. Once an ADAS is installed on a car, it will have been trained on a large number of drivers, but there is no guarantee that it will work well with a completely new subject without re-training. In this work we tackle this problem by applying the domain-adversarial training technique to encourage a Recurrent Neural Network to adapt to a smaller, unlabelled sets of driving data in which the features may have different marginal distributions. To obtain meaningful results we also decided to design an improved model for manoeuvre anticipation, addressing some of the shortcomings of previous work. In this section we provide a brief survey on the most recent research on action prediction and manoeuvre anticipation, as well as on the latest techniques in the area of domain adaptation, with special focus on those applied to deep learning and sequential data. For an in-depth review of the general field of domain adaptation, we suggest the papers by Jiang \citep{Jiang2008b} and Patel et al. \citep{Patel2015b}

	\subsection{Action Prediction}
	Predicting future actions differs from simple classification tasks, in that the aim is to anticipate an event with as little data as possible. One way to do this practically is by setting a threshold, processing the data at each time-step of an input sequence of observations, and only making a prediction when the probability of the corresponding output is above that threshold. To do so, a network needs to be able to remember past states and observations within a single time-series, in order to learn the temporal relations between latent features in that sequence. Most recent examples of action anticipation models do this by using LSTM- or GRU-based architectures \citep{Gamboa2017,Liu2016Spatio-temporalRecognition,Greff2016,Baccouche2011}. LSTMs and GRUs are specific types of RNN units which are able to remember and forget previous states through a number of modulating gates \citep{Hochreiter:1997:LSM:1246443.1246450,Chung2014EmpiricalModeling}. They have been proven to solve the problem of vanishing gradients, which affects heavily "vanilla" recurrent networks, and thus are able to perform very well with long time series \citep{Gers2000}. This aspect makes them more suitable to process sequential observations than regressive models such as Gaussian Mixtures and non-recurrent Neural Networks, or Hidden Markov Models (HMMs), which assume that each observation's probability only depends on the current state and are thus not suitable for modelling contextual effects and long sequences \citep{Ravichandar2016b}. Despite these shortcomings, HMMs have been shown to produce promising results, especially compared to approaches that do not employ deep learning \cite{Jain2016}; they could therefore provide an excellent alternative in cases where the high computational requirements of RNNs cannot be met, for instance in mobile applications. A further improvement on the HMM approach is represented by Markov Decision Processes, which have been shown to provide reliable and accurate predictions for long-term driving risk inference \citep{shimosaka2015risk}.\\
	In order to make predictions within a few time-steps, rather than simply classifying an action when the whole sequence has been processed, a recent trend has been to implement custom loss functions which exponentially increase with time. Later classification mistakes are penalised more heavily than earlier ones: the model is thus encouraged to provide a confident prediction as soon as possible. In multi-class tasks such as manoeuvre anticipation, these functions are often modifications to the standard cross-entropy loss. Aliakbarian et al. \citep{Aliakbarian2017b} included this type of time-dependent loss in a multi-stage LSTM architecture that manages to predict actions accurately with only a small percentage of video sequences, by learning context- and action-related features independently. Chan et al. \citep{Chan2017a} used the exponential loss only for positive examples in a binary classification task to predict driving accidents, also including it in a LSTM-based RNN.
	\subsection{Manoeuvre Anticipation}
	Models for manoeuvre anticipation are similar to those for action prediction; the differences are the specificity of the subject's movements (mostly head and eyes) and the type of contextual information obtained from the environment. At the time of writing, the most recent and possibly complete approach to anticipate driving manoeuvres has been proposed by Jain et al. \citep{Jain2016} (Brain4Cars). Their model implements an LSTM-based architecture with high-level sensory fusion to process multi-modal observations, using facial landmarks, head pose, car speed, GPS information and lane configuration. An exponential loss function was also included, which was proven not only to successfully encourage early predictions, but also to act as a regulariser. In addition, they provided a complete dataset of driving videos, showing the driver and the outside environment in the few seconds before turns and lane changes. Because of Brain4Cars' promising results, in this work we used their architecture and dataset respectively as benchmark and for evaluation purposes. \\
	An area of improvement identified by the Brain4Cars team lies in the addition of eye tracking information to the model's inputs. Gaze direction, in fact, has been shown to generally correlate with the direction of the subsequent movement \citep{Flanagan2003,Gredeback2015,Matsumoto199,Miyajima2016b}. Including gaze direction as an additional feature vector is particularly important in the context of manoeuvre anticipation: eye movements to look in the mirrors or at objects on the road may not necessarily be accompanied by a movement of the head, but may provide information about the direction of a consequent manoeuvre. It has been argued that the choice not to implement it is justified by the difficulty of obtaining accurate measurements of gaze direction without using specialised hardware \citep{Jain2016}. Fletcher and Zelinksy \citep{Fletcher2009a}, for instance, successfully implemented an ADAS which analyzes driver inattentiveness using gaze tracking, and Ravichandar et al. \citep{Ravichandar2016b} used prior probabilities based on the eye gaze to enhance the accuracy of their action-prediction model; both works were carried out using ad-hoc tracking cameras. However, recent studies have shown that it is in fact possible to estimate gaze direction accurately from videos captured by regular high definition cameras \citep{Gou2017a,Baltrusaitis2016OpenFace:Toolkit}. \\
	While it is clear that eye tracking may provide additional benefits to infer the intentions of a driver, adding additional sensors to the model requires optimal sensory fusion in order for deep networks to perform well \citep{Ngiam2011}. Indeed, a share of the recent literature has focused on improving the integration of the multi-modal data coming from both inside and outside of the car. For instance, Doshi et al. \citep{Doshi2011b} used a relevant vector machine (RVM) model to detect lane-change intent; they exploit the ability of the RVM to obtain a sparse data representation of the dataset, thus performing sensory fusion by automatically choosing discriminating features from multi-modal signals. Tawari et al. \citep{Tawari2014Looking-inBraking} developed a "Merge and Lane Change Assist" system in which sensory fusion is achieved by encoding all constraints (spatial, temporal, as well as legal) into a compact probabilistic representation \citep{Sivaraman2014}. Jain et al. \citep{Jain2016} used a solution similar to those proposed by Sung et al. \citep{Sung2015a} as well as Yang and Eisenstein \citep{Yang2017a}, which is not to join the features before feeding them to the network, but rather to concatenate their high-level representations through learnable layers of the neural network. This is a very simple and scalable solution that only requires simple modifications to the architecture of the network.

	\subsection{Domain Adaptation}

	Domain adaptation is a type of transfer learning where two domains (\textit{source} and \textit{target}) share their feature space but have different marginal distributions \citep{Purushotham2016a}. It attempts to solve the issue of enabling a model trained on a certain dataset (source domain) to perform well on a differently distributed dataset, of which the labels are completely or partially unknown \citep{Ben-David2007}. This problem has been investigated in computer vision \citep{Saenko2010,Gong2012,Fernando2013} and natural language processing \citep{Foster2010} using a variety of different approaches to reduce the discrepancy between the two domains, including alignments of the subspaces \citep{Fernando2013}, parameter augmentation \citep{Watanabe2016b}, domain-invariant projection \citep{Baktashmotlagh2013},
	and instance re-weighting \citep{Jiang2007}. Purushotham et al. \citep{Purushotham2016a} point out that these techniques are not able to capture the temporal dependencies in sequential data, and those which do (for instance through a Bayesian approach \citep{Huang2009} or RNNs \citep{Socher2011}), cannot accurately infer non-linear relationship. Deep learning approaches, on the other hand, have proven successful in capturing time dependencies and complex, non-linear, domain-invariant relationships through domain adaptation. Examples include marginalised denoising autoencoders \citep{Chen2012}, ad-hoc CNN architectures \citep{Tzeng2015}, and feature embeddings \citep{Yang2014}.\\
	Out of the deep learning approaches, one of the most elegant and easy-to-implement solutions is the Domain-Adversarial Neural Network (DANN), developed by Ganin et al. \citep{Ganin2016Domain-adversarialNetworks}. DANNs perform domain adaptation by learning domain-invariant features through a neural network architecture composed of three sections: a feature extractor, a discriminative classifier, and an adversarial domain classifier, whose loss is \textit{maximised} through a special gradient reversal layer. The domain classifier's role is to encourage the feature extractor to find latent representations of the features which are domain-invariant \citep{Purushotham2016a}. In particular, this type of domain adaptation aims at creating a common subspace for the source and target domains, so that the trained model can classify examples from the target domain without having access to the labels of the target's training set \citep{Ganin2016Domain-adversarialNetworks}. The ability to adapt in an unsupervised manner is particularly appealing for manoeuvre anticipation models, as it would allow them to generalise well to new drivers simply by retraining them using a set of unlabelled driving videos as the target domain. While this method has been originally carried out using a CNN for the feature extraction stage, Purushotam et al. applied the technique of adversarial training on a RNN (R-DANN) and a variational RNN (VRADA), in order to capture temporal relationships from one domain to the other \citep{Purushotham2016a}. Our paper introduces the same concept applied to manoeuvre anticipation, expanding on the simpler R-DANN architecture and using an LSTM-GRU network as the feature extractor. \\
	Lastly, since finding domain-invariant features in multi-modal time series can be challenging and computationally expensive, it has been suggested that fine-tuning the feature extractor section on the source dataset can yield to an improved performance without sacrificing domain invariance \citep{Long2016,Ganin2016Domain-adversarialNetworks}. This has been shown to be effective even in non-adaptive cases of time-dependent features \citep{Aliakbarian2017b}, and its application to the manoeuvre anticipation problem will also be demonstrated in this paper.

\section{Model for Manoeuvre Anticipation}

	In order to apply the domain-adversarial approach to manoeuvre anticipation, we first propose an improved RNN architecture based on LSTMs and another closely-related type of gated layer, the GRU. Both prevent the gradient of the loss function from either vanishing or exploding during backpropagation thanks to the activation functions of the layers' \textit{gates}, which are learnable and create sums of activations over which the derivatives can distribute  \citep{Jain2016,Goodfellow-et-al-2016}. The gradient can thus propagate for a long time, allowing long time-series to be processed. Before illustrating our model, we provide a brief explanation of these two layers.

	\subsection{LSTM and GRU}

	\begin{figure}[!h]
		\centering
		\begin{subfigure}{.44\textwidth}
			\centering
			\includegraphics[width=\linewidth]{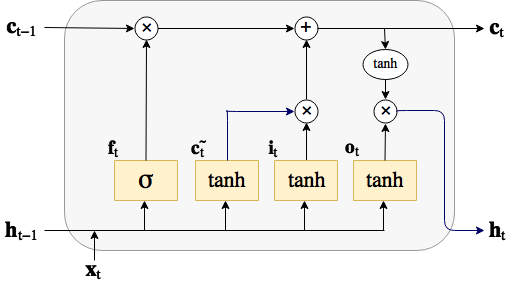}
			\caption{LSTM.}
			\label{fig:lstm}
		\end{subfigure}
		\begin{subfigure}{.50\textwidth}
			\centering \includegraphics[width=\linewidth]{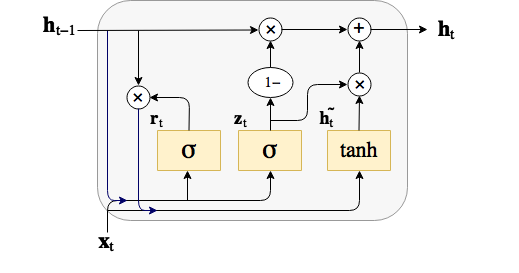}
			\caption{GRU.}
			\label{fig:gru}
		\end{subfigure}
		\caption{\textbf{Diagrams of Long Short-Term Memory (LSTM) and Gated Recurrent Unit (GRU).}}
		\label{fig:lstm_gru}
	\end{figure}

	The structure of a typical LSTM can be seen in Fig.\ref{fig:lstm}. It consists of a memory cell ($\bm{c}$), which allows information to be accumulated over long sequences, and three gates. The forget gate ($\bm{f}$) controls how the information in the memory cell is updated, deleted, or stored; the input gate ($\bm{i}$) takes in current observations and writes new values to the memory cell; and the output gate ($\bm{o}$) computes the hidden output based on the content stored in the memory cell \citep{Goodfellow-et-al-2016}. At every time-step $t$ of a time-series observation, the operations are computed in the following order: the activations of the input ($\bm{i}_t$) and forget gates ($\bm{f}_t$) are calculated, and the memory
	cell is updated ($\bm{c}_t$). Subsequently,
	the output of the cell is finally produced as a hidden representation ($\bm{h}_t$) depending on the activation function of the output gate ($\bm{o}_t$). The inputs into each unit are the observations ($\bm{x}_t$), the previous cell state $\bm{c}_{t-1}$, and
	the output $\bm{h}_{t-1}$ from the LSTM at $t−1$  \citep{Jain2016}.
	The process is defined by the following equations:
	\begin{align}
	&\bm{i}_t = \tanh(\bm{W}_{xi}x_t + \bm{W}_{hi}h_{t-1} + \bm{W}_{ci}\bm{c}_{t-1} + \bm{b}_i) \\
	&\bm{f}_t = \sigma(\bm{W}_{xf}x_t + \bm{W}_{hf}h_{t-1} + \bm{W}_{cf}\bm{c}_{t-1} + \bm{b}_f) \\
	&\bm{o}_t = \tanh(\bm{W}_{xo}x_t + \bm{W}_{ho}h_{t-1} + \bm{W}_{co}\bm{c}_{t-1} + \bm{b}_o) \\
	&\bm{c}_t = \bm{f}_t\; \odot\; \bm{c}_{t-1}\; +
	\bm{i}_t\; \odot\; \tanh(\bm{W}_{xc}\bm{x}_t + \bm{W}_{hc}\bm{h}_{t-1} + \bm{b}_c) \\
	&\bm{h}_t = \bm{o}_t\; \odot\; \tanh(\bm{c}_t)
	\end{align}
	where $\bm{W}_*$ are the weights and $\bm{b}_*$ the biases. $\odot$ is the Hadamart product, also known as element-wise or point-wise vector product. \citep{Jozefowicz2015,Jain2016}. \\
	The GRU, pictured in Fig.\ref{fig:gru}, is similar to the LSTM, but is lacking the memory cell and the output gate \citep{Cho2014}. It is defined by the following equations:
	\begin{align}
	&\bm{r}_t = \sigma(\bm{W}_{xr}x_t + \bm{W}_{hr}h_{t-1} + \bm{b}_r) \\
	&\bm{z}_t = \sigma(\bm{W}_{xz}x_t + \bm{W}_{hz}h_{t-1} + \bm{b}_z) \\
	&\bm{h}_t = \tanh(\bm{W}_{xh}x_t + \bm{W}_{hh}(\bm{r}_t \odot \bm{h}_{t-1}) + \bm{b}_h) \\
	&\bm{\widetilde{h}}_t = \bm{z}_t \odot \bm{h}_{t-1} + (1 - \bm{z}_t) \odot \bm{\widetilde{h}}_t
	\end{align}
	As it can be seen, the number of operations computed at each time-step is lower than for the LSTM. It was also found that GRUs outperform LSTMs in specific situations, for instance when no dropout is used \citep{Chung2015,Jozefowicz2015}. This can be explained by a lower tendency to overfit due to the reduced complexity. While the introduction of dropout makes LSTM better choices in most situations, using GRUs lowers computation and training time by reducing the number of learnable parameters. They thus represent a valuable option.\\
	Throughout this paper, LSTM and GRU operations will be referred to as:
	\begin{align}
	(\bm{h}_t,\bm{c}_t) &= \text{LSTM}(\bm{x}_t, \bm{h}_{t-1}, \bm{c}_{t-1}) \\
	(\bm{h}_t) &= \text{GRU}(\bm{x}_t, \bm{h}_{t-1})
	\end{align}

	\subsection{Anticipation Framework}

	To build our model we followed the framework for manoeuvre anticipation proposed and defined by Jain et al. \citep{Jain2016}, which we summarise here. The inputs to the model at training time are represented by $N$ time series in the form of $\{(\bm{x}_1,..., \bm{x}_T )_i, \bm{y}_i\}^N_{i=1}$, in which $\bm{x}_t$ is the set of features at time-step $t$, and $i$ is the index of the sequence. $\bm{y} = [y^1,...,y^J]$ is the representation of the action occurring at the end of the sequence when $t = T$, with $y_j$ standing for the probability of the sequence leading to event $j$. In our case, $J = 5$: four manoeuvres (turning right, turning left, lane change right, lane change left), plus going straight as the default action.  During training, each manoeuvre is represented by a one-hot-encoded vector. At test time, an observation vector $\bm{x}_t$ is received by the model at every time-step. A probability threshold $p_{th}$ is chosen, so that when, and only when, any $y^j \geq p_{th}$, a prediction of the respective manoeuvre will be made. At $t = T$, if no manoeuvre has been predicted with sufficient confidence, the default action (going straight) will be predicted. \\
	The inputs $\bm{x}$ consist of the matrices of the head features (composed of facial features and head pose), $\bm{\phi} = [\bm{\phi}_1, …, \bm{\phi}_T]$; of the gaze,
	$\bm{\gamma} = [\bm{\gamma}_1, …, \bm{\gamma}_T]$; and of the environmental features,
	$\bm{\eta} = [\bm{\eta}_1,…, \bm{\eta}_T]$. Each element $\bm{\phi}_t$, $\bm{\gamma}_t$, and $\bm{\eta}_t$ is a vector containing the respective individual features. The feature engineering and data processing pipelines are described more in detail in Section \ref{sec:data}.

	\subsection{Network Architecture}
	\label{sec:architecture}

	\begin{figure}[!h]
		\centering
		\includegraphics[width=0.6\linewidth]{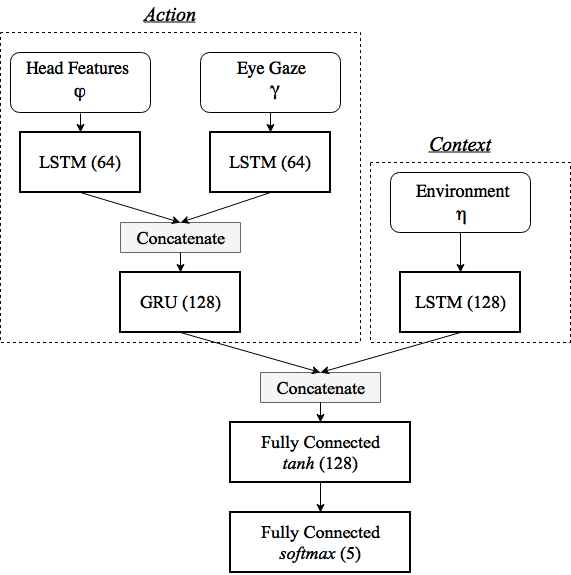}
		\caption{\textbf{Architecture of our model}. Sensory fusion is performed by learning the concatenation of higher-level representations of the features at different points in the network.}
		\label{fig:model}
	\end{figure}

	The structure of our architecture, shown in Fig.\ref{fig:model}, was inspired by the sensory fusion approach proposed in previous work, in which the inputs are not concatenated before going through the network, but rather after the first recurrent layer(s) as high-level features \citep{Jain2016,Sung2015a,Yang2017a}. We enhanced this idea by incorporating the concept of action- and context-dependent features proposed by Aliakbarian et al. \citep{Aliakbarian2017b}. They showed that, for a similar type of anticipation problem, this structure performs better than two parallel recurrent layers which disregard the action-context distinction. The concatenation occurs at two different places in the network: first the representations of the head features $\bm{\phi}_t$ are concatenated with those of gaze $\bm{\gamma}_t$, after each of them has gone through an LSTM layer (Eq.\ref{eq:face} and \ref{eq:gaze}). Their concatenation is then passed through a GRU layer (Eq.\ref{eq:fusion1}). The rationale for this choice is as follows: when an event can be classified in a main type (manoeuvre vs. going straight, i.e. no manoeuvre) and a sub-type (each of the four manoeuvres), a two-layered RNN architecture enables each of the two layers to specialise in the two predictions. This idea was proposed and successfully tested by Xiao et al. \citep{Xiao2017d}. In our case, the first LSTM learns the main event type, while the following GRU learns which manoeuvre is performed. A GRU layer was preferred to a second LSTM in order to limit the increase in complexity. \\
	The output of the GRU layer is then concatenated with the hidden output of a third LSTM layer whose inputs are the environmental features $\bm{\eta}_t$ (Eq.\ref{eq:context}). This concatenation is then fed to a fully-connected dense layer (Eq.\ref{eq:fusion2}) \citep{Jain2016}. Finally, the last softmax layer provides the classification probabilities for the five classes (Eq.\ref{eq:softmax}). \\
	\begin{align}
	(\bm{h}_t^\phi,\bm{c}_t^\phi) &= \text{LSTM}_\phi(\bm{\phi}_t,\bm{h}_{t-1}^\phi,\bm{c}^\phi_{t-1}) \label{eq:face} \\
	(\bm{h}_t^\gamma,\bm{c}_t^\gamma) &= \text{LSTM}_\gamma(\bm{\gamma}_t,\bm{h}_{t-1}^\gamma,\bm{c}^\gamma_{t-1}) \label{eq:gaze}\\
	(\bm{h}_t^a) &= \text{GRU}_a([\bm{h}_t^x;\bm{h}_t^g],\bm{h}_{t-1}^a,\bm{c}^a_{t-1}) \label{eq:fusion1}\\
	(\bm{h}_t^\eta,\bm{c}_t^\eta) &= \text{LSTM}_\eta(\bm{\eta}_t,\bm{h}_{t-1}^\eta,\bm{c}^\eta_{t-1}) \label{eq:context}\\
	\bm{z}_t &= \text{tanh}(\bm{W}_f[\bm{h}_t^a;\bm{h}_t^\eta] + \bm{b}_f) \label{eq:fusion2}\\
	\bm{y}_t &= \text{softmax}(\bm{W}_y \bm{z}_t + \bm{b}_y) \label{eq:softmax}
	\end{align}

	Because of the high number of learnable parameters ($>18\times10^4$), dropout was applied to both the recurrent and dense layers of the network. A dropout of 0.6 was set to the LSTM and GRU recurrent connections (\textit{recurrent} dropout), while a dropout of 0.7 was set to the output of every layer \citep{Gal2010,Bluche2015}. Additionally, the bias of the recurrent layers was initialised to 1, in order to improve the model's performance and training \citep{Jozefowicz2015,Gers2000}.\\
	In order to encourage the model to anticipate early, we implemented the loss function shown in Eq.\ref{eq:loss}, which weighs each term of the cross-entropy categorical loss with an exponential term, as a function of time. The loss will be greater at larger values of $t$, thus rewarding the network for predicting the right class as early as possible \citep{Aliakbarian2017b,Chan2017a,Jain2016}. $y_t^j$ represents the probability of event $j$ computed by the model at time-step $t$. \\
	\begin{equation}
	L_y = \sum\limits_{j=1}^N \sum\limits_{t=1}^T -e^{-0.9(T-t)}log(y_t^j) \\
	\label{eq:loss}
	\end{equation}
	The network was trained through backpropagation with gradient-based optimization; we chose the Adam optimiser \citep{Kingma2014Adam:Optimization} for its simplicity, robustness, and low computational cost \citep{Ruder2016AnAlgorithms}. The learning rate was set to $1\times10^{-3}$; the other parameters were set to the values suggested in the original paper: $\beta_1=0.9$, $\beta_2=0.999$, $\epsilon=1\times10^{-8}$, without temporal decay. We implemented additional regularization by using early stopping, with a patience of 80 epochs.

\section{Domain-Adversarial RNN}
	\label{sec:da}

	\begin{figure*}[!h]
		\centering
		\includegraphics[width=0.8\textwidth]{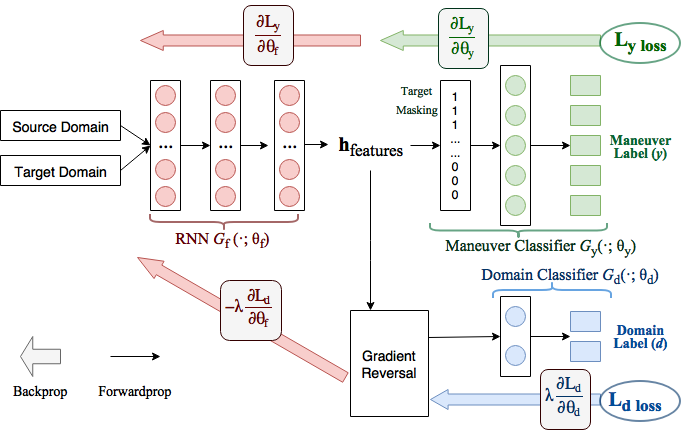}
		\caption{\textbf{DA-RNN}. The proposed architecture for domain-adversarial training of our model to achieve domain adaptation. \textit{Left} (red): the feature extractor, consisting in the LSTM-GRU network show in in Fig.\ref{fig:model}, without the last softmax layer. \textit{Top right} (green): manoeuvre classifier. \textit{Bottom right} (blue): Domain classifier. Diagram style inspired by Ganin et al. \citep{Ganin2016Domain-adversarialNetworks}. (Better viewed in colour.)}
		\label{fig:domain_adversarial}
	\end{figure*}

	Our DA-RNN, pictured in Fig.\ref{fig:domain_adversarial}, is made up of three sections, divided in two main branches. The first section employs the LSTM-GRU network described in Section \ref{sec:architecture} (except for the last softmax layer) as a feature extractor. Its role is to learn the latent relationships between the features in the observation sequences. After the latent features are extracted, their hidden representation is fed into two branches: the first one is a discriminative classifier for manoeuvre anticipation, composed of a single softmax layer; the second one is an adversarial domain classifier. The key part of the latter is the \textit{gradient reversal} layer: during the forward pass, the input is left unchanged, while during backpropagation, the gradient is negated and multiplied by a constant, $\lambda$ (lower section of Fig.\ref{fig:domain_adversarial}). The loss of the domain classifier is thus maximised, thereby encouraging the feature extractor to find representations of the features which are domain-invariant \citep{Purushotham2016a}. Conversely, the label classifier's loss is minimised in order for the features to also be discriminative of the manoeuvres. The manoeuvre loss $L_y$ is the time-dependent exponential function in Eq.\ref{eq:loss}, while the domain loss $L_d$ is the regular binomial cross-entropy loss function. The total loss is given by Eq.\ref{eq:tot_loss}; the effect of $\lambda$ is discussed more in detail in Section \ref{sec:experiments}.
	\begin{equation}
	L_{tot} = L_y -\lambda L_d
	\label{eq:tot_loss}
	\end{equation}
	To train the model, half of each input batch is filled with samples from the source domain, and half from the target domain \citep{Ganin2016Domain-adversarialNetworks}. Since the latter, by assumption, contains fewer observations, target samples are used more than once in each epoch. This should not unfairly help the manoeuvre classifier, since $L_y$ is only dependent on samples from the source domain: the observations belonging to the target domain are "hidden" from the manoeuvre classifier by assigning them a loss weight of $0$ through the boolean mask (represented by the "target masking" block in Fig.\ref{fig:domain_adversarial}). The loss contribution from the samples of the source domain is left unchanged by assigning them a loss weight of $1$. This approach is equivalent to computing $L_y$ first using only transformed samples from the source domain, and then computing $L_d$ using the combined batches. However, our method allows for the weight updates to be computed in a single forward- and back-pass, significantly cutting down training time. At test time, inference is made on samples from the target domain not included in the training set, removing the boolean mask from the model. The performance is evaluated only on the prediction of the manoeuvre classifier.

\section{Data Collection and Processing}
	\label{sec:data}

	\subsection{Brain4Cars Dataset}

	The dataset used as a benchmark for testing our anticipation architecture was developed by Brain4Cars \citep{Jain2016}. At the time of writing and to our knowledge, it is the most complete dataset which includes synchronized recordings of the driver's upper body and the road in front of the car. It consists of 700 observations, each including a pair of videos with a duration of 5 seconds: one showing the driver’s face inside the car, and the other one the road ahead, outside of the car. The videos are recorded at 30 frames-per-seconds, for a total 150 frames per video. Additional data is provided for each frame, including lane configuration, speed, and presence of intersections ahead of the car. Every observation is associated with a manoeuvre, performed at the end of the 5 seconds. The numbers of observations for each manoeuvre is as follows: \{Going straight = 234, Changing Lane Left = 124, Changing Lane Right = 58, Turning Left = 123, Turning Right = 55\}. The format of the features in the dataset is similar to that illustrated in Section 5.3 and 5.4. For a more in-depth explanation, we refer the reader to the original work by Jain et al. \citep{Jain2016}. The dataset is publicly available on the Brain4Cars website.\footnote{https://brain4cars.com}

	\subsection{Our Dataset}
\begin{figure*}[!t]
		\centering
		\includegraphics[width=0.8\textwidth]{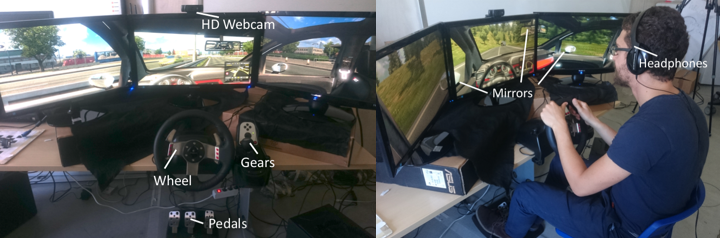}
		\caption{\textbf{The setup for our data collection process.}}
		\label{fig:setup}
	\end{figure*}

	In order to build our own dataset for a more extensive evaluation of our model and the domain adaptation approach, we created a driving simulator set-up following a standard structure for research \citep{Wang2016b,Bergamasco2005} and the same video format of the Brain4Cars data. While the Brain4Cars dataset represents an excellent resource, we believed it necessary to have access to a dataset where \textit{both} the drivers and driving conditions are different from the Brain4Cars dataset in order to evaluate the domain-adaptation capabilities of our model. In practice, the goal is to mimic a real-life situation in which a commercial product may be applied to vehicles that are substantially different in size and/or position of mirrors, conducted by unseen drivers. A driving simulator was chosen in order to collect data quickly and in a controlled environment, using a combination of highway and city driving conditions. Similar virtual simulation setups have been frequently used for validation of ADAS and statistical approaches to driving style predictions \citep{hwang2017validation, Han2019statistical, Wang2017humancentered}.\\
	We used the commercial game Euro Truck Simulator 2 \footnote{SCS Software, 2012, https://eurotrucksimulator2.com/}, including an unofficial modification which allows the player to drive a car instead of a truck. This specific game has been chosen among other driving simulator software due to the realism of the physics engine, the quality of the graphical output on the main view and rear mirrors, as well as the compatibility with the instrumentation. The setup, pictured in Fig.\ref{fig:setup}, consisted of a three-screen high-resolution display system (CPU Intel i7-3930K and GPU NVidia GTX 960) and driving equipment  with pedals, gears, and a steering wheel with force-feedback (G27, Logitech, CH). A commercial web-cam (HD Pro C920, Logitech, CH, at resolution 1080p), placed on the middle screen, was pointed towards the face of the driver. The three-screen setup allowed to have a field-of-view of about $200^{\circ}$. Car data from the game engine was obtained using an unofficial telemetry server \footnote{https://github.com/Funbit/ets2-telemetry-server}, whose output are $JSON$ blocks containing a large number of information such as orientation, speed, acceleration, steering angle, etc. \\
	Five (5) different subjects were asked to play for a time between 30 and 50 minutes, driving both in cities and on highways. They were asked to respect road rules and to behave as if they were in a real car (looking at mirrors, turning their head when needed, etc.) The game screen and the driver's face were recorded synchronously using the Open Broadcaster Studio Open Source software \footnote{https://obsproject.com/}. We aimed at minimizing the variance in the data caused by differences in the camera placement, as well as avoiding time intervals in which the tracking does not capture the facial landmarks due to the camera being partially covered by the driver's hands on the wheel. We therefore positioned our camera always in the same position on top of the central monitor, directly in front of the driver seat.\\
	In order to match the Brain4Cars dataset format, we extracted 5-second videos of the desired manoeuvres using the Boris Open Source video annotation tool\footnote{http://www.boris.unito.it/} \citep{Friard2016} to annotate the manoeuvres in the raw footage. A custom Python script was written to cut up the 5 seconds preceding the onset of each manoeuvre (defined as the moment when the wheel touches the lane markings or when it starts turning at the intersection \citep{Jain2016}), using a combination of Open Source OpenCV library\footnote{http://opencv.org/} and FFmpeg video conversion tool\footnote{https://www.ffmpeg.org/}.	The final dataset comprises 113 videos: \{Going straight = 32, Changing Lane Left = 21, Changing Lane Right = 19, Turning Left = 24, Turning Right = 17\}. The number of observations is smaller than those in the Brain4Cars dataset, as its aim is to provide the target samples for a domain adaptation problem; in a real-life situation, target sets will always be considerably smaller than source sets.

    Our dataset has been made available as Open Access on Zenodo with the following DOI: \url{http://dx.doi.org/10.5281/zenodo.1009540}.

	\subsection{Feature Extraction and Processing}

	To extract features from the videos of both datasets we used OpenFace, an open-source toolkit that provides facial landmark tracking, head pose estimation, and gaze tracking, from videos and images, using a combination of Conditional Local Neural Fields (CLNF) and CNNs \citep{Baltrusaitis2016OpenFace:Toolkit}. It is capable of high performance real-time tracking using regular cameras, making it an attractive option for ADAS. Once the data was extracted from the videos, we followed the feature processing pipeline defined by Jain et al. \citep{Jain2016}, with slight modifications to binning intervals, labelling of the environmental features, and feature scaling. We were interested in obtaining the movements of the facial landmarks, the head pose, the direction of the gaze, as well as environmental informations – namely lane configuration, presence of intersections ahead in near proximity, and speed of the car. A time-series is constructed for each sample in the datasets, with each frame representing a time-step in the observation sequence $\bm{x}_t$.
	Data analysis and processing, feature engineering were carried out in Python 3.

	\subsection{Action-related Features}
	\subsubsection{Facial Landmarks and Head Pose}
	We represent the movement of the driver's head with the motion of the facial landmarks, using a binning approach. We took the velocity of each of the 68 landmarks between consecutive frames, calculating the horizontal motion as $\delta x^{face} = x^{face}_t - x^{face}_{t-1}$, in pixels and in the image space; and the angular motion in the $x-y$ plane as $\theta^{face} = arctan2(\delta y^{face}, \delta x^{face})$, in radians. These values were binned to create histogram features. Six (6) bins were chosen for the horizontal motion:\\
	$\{\delta x\leq−5; -5\textless\delta x\leq-2.5; −2.5\textless\delta x\leq0; 0\textless\delta x\leq2.5; 2.5\textless\delta x \leq5; \delta x\textgreater 5\}$ pixels, while for the angular motion we used four (4) bins: \\
	$\{0 \textless \theta \leq \pi/2;\; \pi/2 \textless \theta \leq \pi;\; \pi \textless \theta \leq 3\pi/2;\; 3\pi/2 \textless \theta \leq 2\pi\}$ radians.\\ Negative values refer to motions towards the left-hand side of the image and the right-hand side of the driver.\\
	The head pose was also included using the Euler angles representation [$R_x(\alpha_{pitch})$, $R_y(\alpha_{yaw})$, $R_z(\alpha_{roll})$] as it was shown to improve the performance \citep{Jain2016,Trivedi2005}. Overall, the histogram features and head pose form the head features $\bm{\phi} \in \mathbb{R}^{13}$.

	\subsubsection{Eye Gaze}
	Information about the driver's gaze was also obtained through OpenFace's output, which provides a 3D direction vector for each eye, normalised and in world coordinates. We took the average components of both eyes, and applied a Butterworth low-pass filter (4th order, sampling frequency of \SI{30}{\hertz}, cutoff of \SI{1.66}{\hertz}) in order to eliminate noise deriving from tracking inaccuracies and natural saccades. The filter introduces a minimal delay which was proven to not be detrimental to the model, and it can be applied in real time. To create the feature vector we take the $x$ and $y$ component of the gaze direction vector (which correspond to the horizontal and vertical direction in the image plane), scaled between -1 and 1, to create histogram features, similarly to the head features. The bins chosen are:\\ $\{−1\textless\delta x\leq−0.5; -0.5\textless\delta x\leq0; 0\textless\delta x\leq0.5; 0.5\textless\delta x\leq1\}$ pixels, in the image space. Identical bins were chosen for $\delta y$. In this case we used the direction components rather than the inter-frame velocity since we found that it correlates better with the driver's intention. We define $\bm{\gamma} \in \mathbb{R}^8$ as the gaze features.

	\subsection{Context-related Features (Environment)}
	The environmental context is expressed by $\bm{\eta} \in \mathbb{R}^4$. The first two features are boolean variables expressing the presence of a lane to the left and to the right of the car, respectively. The third feature is also boolean, and indicates the presence of an intersection ahead and in the near proximity of the car. For the purpose of this study, these three values were labelled manually for each observation. In a practical implementation, this information could be extracted automatically through lane detection algorithms \citep{Kaur2015LaneReview,Kim2008} fused with GPS data. Finally, the fourth value is the speed of the car in \SI{}{\km/\hour}. This value is provided in the Brain4Cars dataset, and was measured by the physics engine of the game in our dataset.
	\section{Experiments}
	\label{sec:experiments}
	In this section we present the results of the experiments conducted on the Brain4Cars' and our dataset, using the features extracted according to the method described in Section \ref{sec:data}. We first tested the network for manoeuvre anticipation on the two datasets separately, shuffling samples from all drivers before the training-test split, as done by Jain et al. \citep{Jain2016}. We show that the new architecture and additional features yield to an improvement in the performance of the model. We then investigated the applications of domain adaptation to our model via domain-adversarial training, in two ways: a) employing a leave-one-out approach on each driver in the Brain4Cars dataset, in order to study the possibility of personalising a model in a partly-unsupervised manner;
	b) using the entirety of the Brain4Cars dataset as the source domain and our dataset as the target domain, in order to study how the approach works with different feature distributions. In both cases, we attempted to increase the performance of the models through fine-tuning –meaning initializing the weights of certain layers with weights from a pre-trained model–, and we studied the effect of varying the value of the domain loss multiplier $\lambda$. \\
	To evaluate the results, four measures were used. Three of them provide the multi-class classification score for the predicted action, namely: precision, recall and F1 score \citep{sokolova2009systematic}. The fourth measure is the time-to-prediction (TTP), representing how many seconds before the onset of the manoeuvre the prediction is made. When calculating precision, recall, and F1 score, the "going straight" predictions are not considered, as it is considered the baseline state.
	The probability threshold was set to 0.9 for all experiments, since this value was shown to yield the most confident and quick predictions \citep{Jain2016}. This means that at each time-step $t$, a prediction is made only if at least one of the 5 outputs of the models is $\geq 0.9$. We trained all models with batches of size 128.

	\subsection{Experiment 1: Manoeuvre Anticipation}

	The first experiment was performed using the LSTM-GRU architecture described in Section \ref{sec:architecture} on the Brain4Cars dataset, without domain adaptation, in order to evaluate its performance as a feature extractor. Samples from all drivers were shuffled, and a test set was set aside taking 15\% of the total observations. 5-fold cross validation was used, with the 5 validation subsets each making up a different 20\% of the training set. Training and validation sets were normalised jointly before the split. Following Jain et al.'s approach, we augmented the training set by extracting subsequences of random length ($T_{sub}$) from the original observation, with $50 < T_{sub} < 150$. These additional samples were used as additional training examples, thereby adding redundancy and additionally decreasing the risk of over-fitting \citep{Jain2016}. More sub-samples were taken from under-represented classes, thus balancing the class ratios. For the Brain4Cars dataset, this led to a total of 2160 training samples; 312 for ours.\\
	The results are shown in Table \ref{tab:performances}. We compare our performance with Jain et al.'s LSTM model \citep{Jain2016}. It can be seen that our network performs better for all manoeuvres, reaching a higher F1 score as well as a higher time-to-prediction.
    We theorise that this improvement is due partly to the additional information given by the gaze direction. Indeed, excluding the gaze-related features, the performance of the model was observed to drop by up to 1-1.5\%. However, we attribute the improvement especially to the enhanced sensory fusion (including the action-context structure). Indeed, a parallel structure in which the hidden representations of all three inputs are concatenated after the initial recurrent layers performs marginally worse than Brain4Cars' model. This shows that simply adding more features does not necessarily improve the performance of a model if it is not accompanied by rational modifications in the architecture. \\
	In addition, it was noticed that the added complexity of the network caused a strong tendency to overfit, despite the high dropout, the early-stopping criteria, and the redundancy in the training data. However, it is likely that the number of samples is not large enough; bigger datasets should alleviate this problem and further enhance the model's performance. Lastly, we report that including the car's speed as part of the environmental features was detrimental to the performance of the model, and, for this reason, it was excluded in the subsequent experiments. This may be due to the fact that, especially in sequences leading to lane changes or driving straight, maneuvers may not be correlated with specific ranges of speed values; further analysis should be carried out on the subject. However, we observed that excluding any of the other features used by Jain et al. \citep{Jain2016} from either dataset worsens the performance considerably. This was confirmed experimentally through 5-fold cross-validation.

	\begin{table}[!h]
		\centering
		\resizebox{\linewidth}{!}{\begin{tabular}{|M{3.3cm}||M{5.3cm}|M{5.3cm}|M{5.3cm}||}
				\hline
				& \begin{tabular}{@{}c@{}} Changing lane \end{tabular} & \begin{tabular}{@{}c@{}} Turning \end{tabular} &  \begin{tabular}{@{}c@{}} All manoeuvres \end{tabular} \\ [0.4ex]
				\hline
				\addlinespace[0.5ex] & \begin{tabular}{M{0.8cm} M{0.8cm} M{0.8cm} M{0.8cm}}  Prec. (\%)  & Recall (\%) & F1 (\%)  & TTP (s) \end{tabular}
				& \begin{tabular}{M{0.8cm} M{0.8cm} M{0.8cm} M{0.8cm}}  Prec. (\%) & Recall (\%) & F1 (\%) & TTP (s) \end{tabular}  &
				\begin{tabular}{M{0.8cm} M{0.8cm} M{0.8cm} M{0.8cm}}  Prec. (\%) & Recall (\%) & F1 (\%) & TTP (s) \end{tabular} \\ [0.4ex]
				B4C &  \begin{tabular}{M{0.8cm} M{0.8cm} M{0.8cm} M{0.8cm}}  95.4 & 85.7 & 88.8 & 3.42 \end{tabular}  & \begin{tabular}{M{0.8cm} M{0.8cm} M{0.8cm} M{0.8cm}}  68.5& 78.5  & 72.1 & 3.78 \end{tabular} & \begin{tabular}{M{0.8cm} M{0.8cm} M{0.8cm} M{0.8cm}}  82.0  &  82.1  & 82.0 & 3.58 \end{tabular} \\
				B4C w/ Head pose &  \begin{tabular}{M{0.8cm} M{0.8cm} M{0.8cm} M{0.8cm}} / & / & / & / \end{tabular}  & \begin{tabular}{M{0.8cm} M{0.8cm} M{0.8cm} M{0.8cm}} / & /  & / & / \end{tabular} & \begin{tabular}{M{0.8cm} M{0.8cm} M{0.8cm} M{0.8cm}}  90.5  &  87.4  & 88.3 & 3.16 \end{tabular} \\
				LSTM-GRU (ours) &  \begin{tabular}{M{0.8cm} M{0.8cm} M{0.8cm} M{0.8cm}} 96.5 & 90.5 & 93.6 & 3.90 \end{tabular}  & \begin{tabular}{M{0.8cm} M{0.8cm} M{0.8cm} M{0.8cm}}  91.1 & 90.9 & 91.0 & 4.06 \end{tabular} & \begin{tabular}{M{0.8cm} M{0.8cm} M{0.8cm} M{0.8cm}}  \textbf{92.3} &  \textbf{90.8} & \textbf{91.3} & \textbf{3.98} \end{tabular} \\ [0.2ex]
				\hline
		\end{tabular}}
		\caption{\textbf{Manoeuvre anticipation on the Brain4Cars dataset.} Results of the non-adaptive LSTM-GRU network on a test set comprised of observations from all drivers. Results reported from Brain4Cars (B4C) are taken directly from the paper \citep{Jain2016}.}
		\label{tab:performances}
	\end{table}

	The same model was also tested on our dataset, using a similar procedure. The results, illustrated in Table \ref{tab:anticipation_percro}, show that the performance is lower than for the Brain4Cars dataset. This is most likely due to the lower number of samples, and the fact that in a simulation setting, drivers tend to do less emphatic head and eye movements. Conversely, fine-tuning the network by pre-training its weights using the Brain4Cars dataset improves the performance.
	\begin{table}[!h]
		\centering
		\resizebox{0.65\linewidth}{!}{\begin{tabular}{|M{3.5cm}||M{5.5cm}||}
				\hline
				\addlinespace[0.5ex] & \begin{tabular}{M{0.8cm} M{0.8cm} M{0.8cm} M{0.8cm}}  Prec. (\%)  & Recall (\%) & F1 (\%)  & TTP (s) \end{tabular}
				\\ [0.4ex]
				No Pre-Training & \begin{tabular}{M{0.8cm} M{0.8cm} M{0.8cm} M{0.8cm}}  82.0  &  82.1  & 82.0 & 3.9 \end{tabular} \\
				Pre-trained on B4C & \begin{tabular}{M{0.8cm} M{0.8cm} M{0.8cm} M{0.8cm}}  \textbf{89.4} &  \textbf{92.2} & \textbf{90.8} & \textbf{4.1} \end{tabular} \\ [0.2ex]
				\hline
		\end{tabular}}
		\caption{\textbf{Manoeuvre anticipation on our dataset}. Results of the non-adaptive LSTM-GRU network on a test set comprised of observations from all drivers, without and with fine-tuning.}
		\label{tab:anticipation_percro}
	\end{table}
	\FloatBarrier

    Overall, these results show that our model represents an improvement over the most recent LSTM-based approach in manoeuvre anticipation, and prove that our dataset can be used reliably with a comparable performance. Moreover, it was shown that fine-tuning can be a powerful approach to enhance the model's capability if a larger, more reliable dataset is available.

	\subsection{Experiment 2: Domain Adaptation on Different Drivers}

	In order to investigate how a trained model can learn to adapt to a small set of unlabelled driving videos belonging to a subject not included in the training set, we manually separated the videos of each individual driver in the Brain4Cars dataset from each other. We then ran our DA-RNN using observations from all drivers except one as the source domain, while, as the target domain, we used the samples from the remaining driver. Once trained, the model was tested on a set of samples from the target domain not included in the training set, all from the same driver. This was done for each driver separately, and the results were averaged. In addition, we fine-tuned the networks by initializing the weights of the LSTM-GRU feature extractor by pre-training them on the source domain. The results are shown in Table \ref{tab:da1}.
	\begin{table}[!h]
		\centering
		\resizebox{0.7\linewidth}{!}{\begin{tabular}{|M{4.5cm}||M{5.5cm}||}
				\hline
				\addlinespace[0.5ex] & \begin{tabular}{M{0.8cm} M{0.8cm} M{0.8cm} M{0.8cm}} Prec. (\%)  & Recall (\%) & F1 (\%)  & TTP (s) \end{tabular}
				\\ [0.4ex]
				No adaptation & \begin{tabular}{M{0.8cm} M{0.8cm} M{0.8cm} M{0.8cm}}  60.9  &  57.7  & 58.0 & 3.8 \end{tabular} \\
				DA-RNN & \begin{tabular}{M{0.8cm} M{0.8cm} M{0.8cm} M{0.8cm}}  71.7 &  66.5 & 68.1 & \textbf{3.9} \end{tabular} \\
				DA-RNN w/ Fine Tuning & \begin{tabular}{M{0.8cm} M{0.8cm} M{0.8cm} M{0.8cm}}  \textbf{77.7} &  \textbf{75.6} & \textbf{76.8} & 3.8 \end{tabular} \\ [0.2ex]
				\hline
		\end{tabular}}
		\caption{\textbf{Performance of the DA-RNN}, using a leave-one-out approach on the Brain4Cars dataset, averaged over 6 drivers. ($\lambda = 1.10$)}
		\label{tab:da1}
	\end{table}
	\FloatBarrier

	From the results it is clear that when the test driver is not part of the training set, the model does not perform well without adaptation. However, when the model is trained with the domain-adversarial training, an absolute improvement of more than 10\% can be observed. A further improvement is obtained by pre-training the weights of the feature extractors on the source domain, yielding an increase in performance of almost 20\% compared to the non-adapting model. This is because, by initializing the weights with values that are known to produce accurate predictions, the network is already trained to find features that are discriminative of the manoeuvres.

	\FloatBarrier
	Additionally, we found that the hyperparameter $\lambda$, which is the constant multiplier of the domain classifier's loss during backpropagation, plays a key role in feature extraction. The higher its value, the higher the influence of the domain classifier loss, meaning a stronger push towards domain invariance in the feature extractor. The network will therefore tend to find features that are shared by the two domains, but which are not necessarily discriminative. A small $\lambda$, on the other hand, will cause the extracted features to be less domain-invariant but more effective to classify the maneuvers in the source domain samples. For our experiments, we found that a value $\approx 1$, meaning an equal weighting of the losses from the two classifiers yielded the best performance.

	\subsection{Experiment 3: Domain Adaptation on Our Dataset}
	\begin{figure}[!b]
		\centering
		\includegraphics[width=0.6\linewidth]{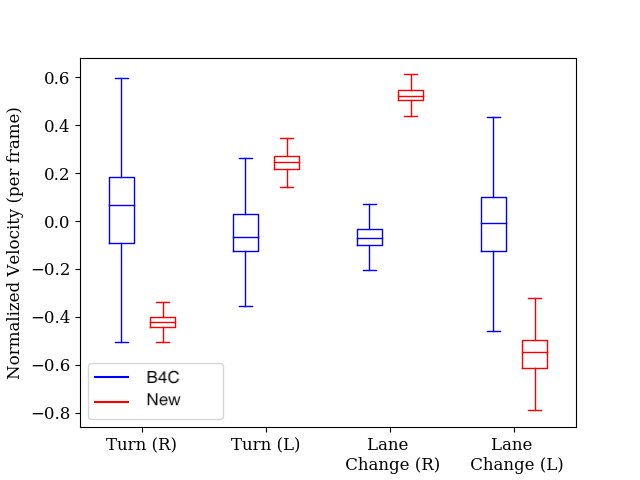}
		\caption{\textbf{Marginal distributions of the Brain4Cars and our datasets.} Box plots of the horizontal velocity of the facial landmarks for both datasets.}
		\label{fig:distr}
	\end{figure}
	The third experiment was performed to attempt a transfer of information between the Brain4Cars data and our dataset. Being able to adapt a model to observations in which the features are distributed very differently can be important in applications such as driver training, Virtual Reality simulations, and autonomous driving research.  Fig.\ref{fig:distr} shows that the two distributions are indeed very different: the head movements made by the drivers in a real-world driving scenario have a higher within-manoeuvre variance than those driving a simulator, but a smaller variance between the different manoeuvres. The fact that the subjects were asked to perform the same movements as in a real-setting, however, makes it plausible that it is possible to find latent features which are shared by both datasets. Moreover, the fact that fine-tuning the model using one dataset improves the performance on the other provides additional support to this theory. Table \ref{tab:da2} illustrates the results of three different approaches:
	\begin{enumerate*}
		\item Training and validating the model on the Brain4Cars dataset, and testing it on our dataset (no adaptation);
		\item Training the DA-RNN using the Brain4Cars dataset as source domain and our dataset as target domain;
		\item Training the DA-RNN using the Brain4Cars dataset as source domain and our dataset as target domain, with fine-tuning (i.e. initializing the weights of the feature extractor by pre-training our LSTM-GRU network on the Brain4Cars dataset.)\\
	\end{enumerate*}

	\begin{table}[!h]
		\centering
		\resizebox{0.7\linewidth}{!}{\begin{tabular}{|M{4.5cm}||M{5.5cm}||}
				\hline
				\addlinespace[0.5ex] & \begin{tabular}{M{0.8cm} M{0.8cm} M{0.8cm} M{0.8cm}}  Prec. (\%)  & Recall (\%) & F1 (\%)  & TTP (s) \end{tabular}
				\\ [0.4ex]
				No adaptation & \begin{tabular}{M{0.8cm} M{0.8cm} M{0.8cm} M{0.8cm}} 27.3 & 31.5 & 29.0 & 3.8 \end{tabular} \\
				DA-RNN & \begin{tabular}{M{0.8cm} M{0.8cm} M{0.8cm} M{0.8cm}} 47.5 & 38.7 & 42.6 & \textbf{4.0} \end{tabular} \\
				DA-RNN w/ Fine Tuning & \begin{tabular}{M{0.8cm} M{0.8cm} M{0.8cm} M{0.8cm}} \textbf{72.6} & \textbf{55.1} & \textbf{62.7} & \textbf{4.0} \end{tabular} \\[0.2ex]
				\hline
		\end{tabular}}
		\caption{\textbf{Performance of the DA-RNN}, with the Brain4Cars dataset as source domain and our dataset as the target domain. ($\lambda = 1.10$)}
		\label{tab:da2}
	\end{table}
	\FloatBarrier
	The first striking observation is the extremely poor performance of the model without adaptation. We infer that when the marginal distribution of the features in two domains are very different, a non-adaptive model fails to generalise. The DA-RNN, in comparison, performs better, with an increase in F1 score of 12 percentage points and an increase in time-to-prediction of \SI{0.2}{\second}. The most critical improvement was registered when the feature extractor was fine-tuned on the source domain, with an F1 score of 62.7\%. The overall performance is still lower than the case of Experiment 2, when the drivers belonged to the same dataset (i.e. driving in similar conditions), but much higher than the no-adaptation case. These relative improvements in performance are comparable to the ones found in recent works performing similar domain adaptation tasks, such as the ones reported in the original DANN paper by Ganin et al. \citep{Ganin2016Domain-adversarialNetworks}. \\ Overall, these results confirm the hypothesis that there exist latent features shared by datasets of observations in which similar driving tasks are performed, but in largely different settings; moreover, they highlight the need of an adaptive approach for the practical implementations and personalizations of ADAS. \\

	The models were created, trained and tested using Python 3, using the deep learning framework Keras 2.0 with the TensorFlow 1.2 backend. Network training was run on a machine with CPU Intel Core i7-7700K and GPU NVidia GTX 1080, and 32GB of memory.

	The code of the architecture of the DA-RNN can be found at \url{https://github.com/michetonu/DA-RNN_manoeuver_anticipation}.

	\section{Conclusions}

	In this paper we proposed and tested a Domain-Adversarial Recurrent Neural Network for adaptive driving manoeuvre anticipation. Trained on a large source dataset of driving observations, our DA-RNN is able to adapt to smaller, unlabelled sets of observations by maximizing the loss of a domain classifier used as an auxiliary output. To extract domain-invariant features from multi-modal time-varying data, we designed a multi-stage LSTM-GRU architecture based on Ganin et al.'s DANN \citep{Ganin2016Domain-adversarialNetworks}, which uses a CNN as feature extractor, and Purushotham et al.'s R-DANN \citep{Purushotham2016a}, which instead uses vanilla RNN layers. In order to apply it to the problem of manoeuver anticipation, we expanded on the work done by Jain et al.\citep{Jain2016}, adding eye gaze direction to the set of input features used to predict driving actions. An alternative approach to carry out advanced sensory fusion is implemented by learning the concatenation of the hidden representations of the features through recurrent layers. The LSTM-GRU section of the network was proven to outperform state-of-the-art work on non-adaptive manoeuvre anticipation tasks. We also present a new dataset obtained through a driving simulation set-up, and made it available for public use. \\
	The evaluation of the DA-RNN was carried out initially using a leave-one-out approach, in which the observations of each individual driver in the Brain4Cars dataset was left out from the training set and used as target domain. An increase of 17 percentage points in F1 score was registered when the DA-RNN's feature extractor was pre-trained on the source domain. The results show the potential of domain-adversarial training to adapt models to new drivers without the need to retrain them with additional labelled examples. In a real-world scenario, observations to be used as the target domain could be captured automatically during the first drive, eliminating the necessity for manual labelling. The second evaluation consisted in using the Brain4Cars dataset as the source domain and our dataset as the target domain, with the same feature space but a different feature distribution. We reported an improvement in F1 score by 114\% (32.7 percentage points) compared to the non-adaptive case, when the model's weights are fine-tuned. Overall, we demonstrate that the domain-adversarial approach represents a promising approach to increase the flexibility and generalization capabilities of commercial ADAS through domain adaptation. \\
	We conclude that non-adaptive models are not able to generalise well in contexts a) where the target driver was not part of the training set, and b) where the features in the target domain have a very different marginal distribution, which is the case when the driving set-up in the target set differs from the one used to collect the training data. Adaptive models will therefore be necessary for ADAS installed in commercial vehicles, and will prove helpful in virtual simulations and driving training tasks. In order to further validate the approach in the context of assisted driving, additional tests should be conducted on a larger target dataset obtained in real-life conditions. Larger and more diverse driving video datasets will enable even higher performances, as one of the bottlenecks of our approach was found in the network's tendency to overfit, due to the complexity of the architecture and the relatively small size of the datasets. This work sets the bases for further research aimed at enabling adaptive deep neural networks to reach performances comparable to fully supervised models.

	\section{Acknowledgments, Contributions and Funding}

	We would like to thank the members of the PERCRO Lab who volunteered for the data collection process, namely L. Peppoloni, G. Dabisias, L. Landolfi, and F. Brizzi. A huge thank you also to J. von Kügelgen for the technical support on domain adaptation and thorough proofreading of the manuscript. Finally, we would like to extend our appreciation towards the creator of all the software used in this work, as well as to the authors of the Brain4Cars dataset for making it available online for free. We strongly believe in open-source software and open access to data; we therefore decided to also publish and share our dataset for free use.

	Study conception and design: MT, ER, CAA. Acquisition of data: MT, AC. Analysis and interpretation of data: MT, ER. Drafting of manuscript: MT, ER, CAA.

	The activities related to the results in this manuscript have been supported within the program of the Excellence Department on Robotics and Artificial Intelligence. The related project is supported by the National Ministry for Education and Research (MIUR). The authors are grateful to Scuola Superiore Sant’Anna, the TeCIP institute and the Department above for the offered logistic, technical and financial support.

	\section{Copyright Notice}

	\textcopyright 2019. This manuscript version is made available under the CC-BY-NC-ND 4.0 license, which can be found at the following URL:\\
	\url{http://creativecommons.org/licenses/by-nc-nd/4.0/}

	\begin{singlespace}
		\section{Bibliography}
		\bibliography{mybibfile}
	\end{singlespace}
\end{document}